\documentclass[runningheads]{llncs}

\usepackage[mobile]{eccv}

\usepackage{eccvabbrv}

\usepackage{graphicx}
\usepackage{wrapfig}
\usepackage{booktabs}
\usepackage{enumitem}
\usepackage{multirow}
\usepackage[accsupp]{axessibility}  

\usepackage{hyperref}

\usepackage{orcidlink}

\begin{document}

\title{CoInteract: Physically-Consistent Human-Object Interaction Video Synthesis via Spatially-Structured Co-Generation}

\titlerunning{CoInteract}

\author{Xiangyang Luo$^{*\dagger}$\inst{1,2} \and
Xiaozhe Xin$^{*\ddagger}$\inst{2} \and Tao Feng\inst{1} \and Xu Guo\inst{1} \and Meiguang Jin\inst{2} \and
Junfeng Ma\inst{2}}

\authorrunning{X. Luo et al.}

\institute{$^1$Tsinghua University, $^2$Alibaba Group \\
\url{https://xinxiaozhe12345.github.io/CoInteract_Project/}}

\maketitle
\let\thefootnote\relax\footnotetext{$^*$\,Equal contribution.\quad $^\dagger$\,Work done during internship at Alibaba Group.\quad $^\ddagger$\,Corresponding author.}

\begin{abstract}
Synthesizing human--object interaction (HOI) videos has broad practical value in e-commerce, digital advertising, and virtual marketing.
However, current diffusion models, despite their photorealistic rendering capability, still frequently fail on (i) the structural stability of sensitive regions such as hands and faces and (ii) physically plausible contact (e.g., avoiding hand--object interpenetration).
We present \textbf{CoInteract}, an end-to-end framework for HOI video synthesis conditioned on a person reference image, a product reference image, text prompts, and speech audio. CoInteract introduces two complementary designs embedded into a Diffusion Transformer (DiT) backbone. First, we propose a Human-Aware Mixture-of-Experts (MoE) that routes tokens to lightweight, region-specialized experts via spatially supervised routing, improving fine-grained structural fidelity with minimal parameter overhead. Second, we propose Spatially-Structured Co-Generation, a dual-stream training paradigm that jointly models an RGB appearance stream and an auxiliary HOI structure stream to inject interaction geometry priors.
During training, the HOI stream attends to RGB tokens and its supervision regularizes shared backbone weights; at inference, the HOI branch is removed for zero-overhead RGB generation.
Experimental results demonstrate that CoInteract significantly outperforms existing methods in structural stability, logical consistency, and interaction realism. 
\end{abstract}

\section{Introduction}
\label{sec:intro}
\begin{figure}[t]
    \centering
    \includegraphics[width=\linewidth]{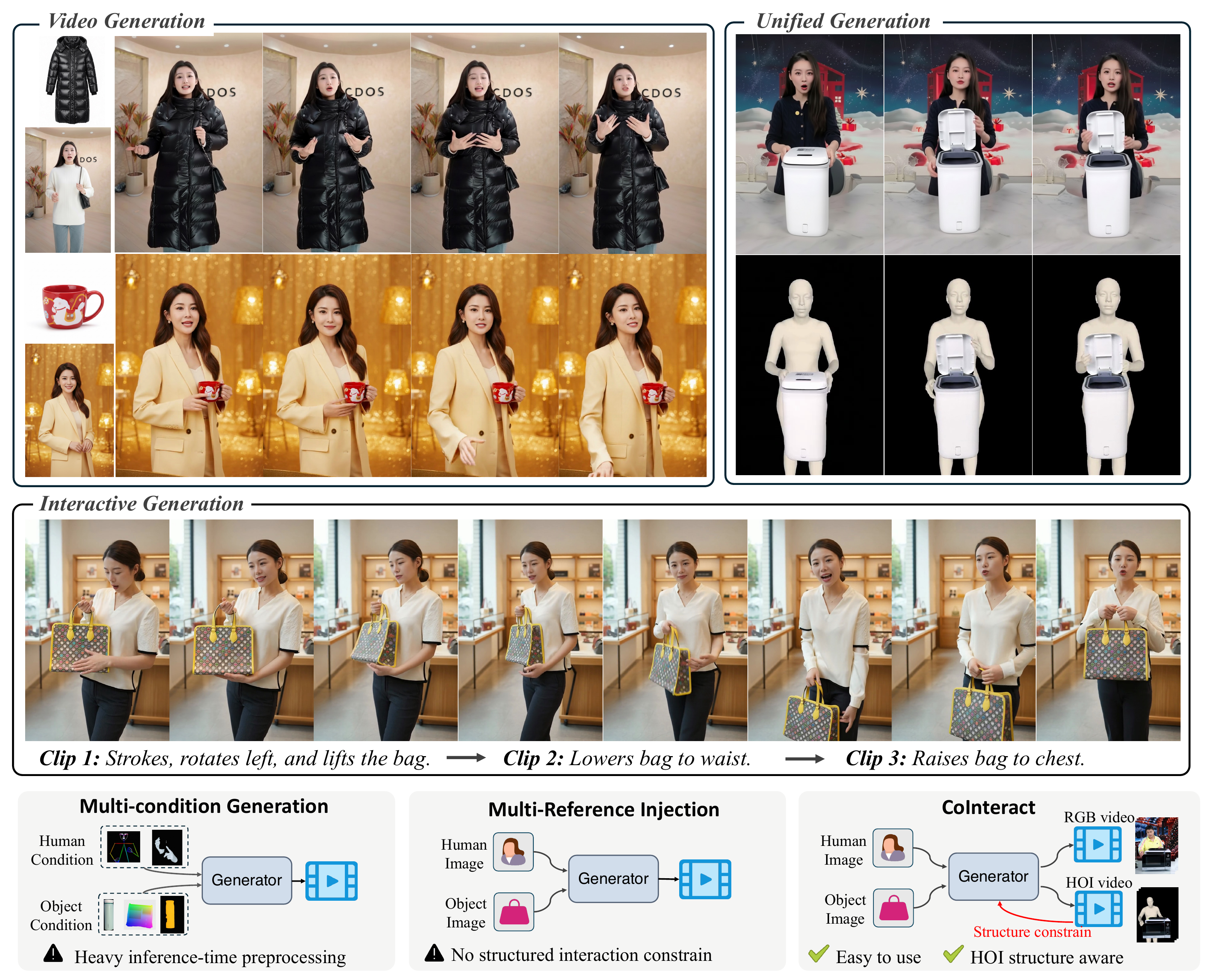}
    \caption{Given two reference images, a text prompt, and audio, CoInteract generates high-fidelity HOI videos. Bottom: Comparison of generation paradigms. Unlike existing methods that rely on heavy inference-time preprocessing and lack structured interaction constraints, CoInteract implements a unified, end-to-end generation framework that is both easy to use and inherently aware of HOI structures.}
    \label{fig:teaser}
\end{figure}
Driven by the remarkable progress in video generation~\cite{wan2025wan,kong2024hunyuanvideo}, speech-driven avatar generation~\cite{xue2025human,zhang2023sadtalker,yu2024gaussiantalker,pointtalk} has achieved unprecedented photorealism, which has opened up vast possibilities for digital humans in e-commerce~\cite{commerce,liao2025humanaesexpert,li2025anydressing}, virtual assistance~\cite{canonswap,guo2026dreamidv}, and remote education~\cite{education,education2}. 
However, as demand shifts from passive speaking to active product demonstration, Human-Object Interaction (HOI) video synthesis has emerged as the next critical frontier, requiring coordinated hand movements, precise object manipulation, and strict physical plausibility beyond what existing talking-avatar generation methods provide.

As shown in Fig.~\ref{fig:teaser}, prior work on HOI video synthesis can be broadly grouped into two paradigms.
\textbf{(i) Multi-condition generation.} These methods extract per-frame human poses and object conditions to guide generation~\cite{xu2026anchorcrafter,liu2025byteloom,wang2025dreamactor}, but the required preprocessing and domain-specific signals often limit robustness and generalization.
\textbf{(ii) Multi-reference generation.} Another line conditions a generative model on person/product references, either by directly synthesizing videos from multiple references~\cite{chen2025humo,jiang2025vace,li2026skyreels} or by compositing references via image editing and then performing speech-to-video generation~\cite{wu2025qwenimage,gao2025wans2v,yang2025infinitetalk}.
While more flexible, these approaches typically lack explicit mechanisms to enforce interaction structure---\eg interaction geometry and physically plausible hand poses---often leading to implausible human--object interaction.

We argue that this limitation is rooted in the RGB-centric nature of current diffusion backbones. As also observed in recent studies~\cite{xue2025mogan,wu2025geometry}, models trained purely on pixel-level supervision have no built-in notion of 3D hand-object spatial relationships or body structure, and must rely on appearance cues alone to infer interaction, which leads to two recurring failure modes: (1) \textit{structural collapse} in hands and faces, where fingers merge or facial features blur, and (2) \textit{physical violations} such as human-object interpenetration, where hands pass through product surfaces due to the lack of explicit object boundary awareness.

To address these challenges, we argue that a model must not only "see" pixels but also "understand" the underlying structural and interaction relationships. In this paper, we present \textbf{CoInteract}, an end-to-end framework that introduces a \textbf{Spatially-Structured Co-Generation} paradigm for physically-consistent HOI video synthesis. Our core philosophy is to embed human structural priors and HOI physical constraints directly into the Diffusion Transformer (DiT) backbone, transforming it from a pure appearance generator into a structure-aware interaction engine.
The technical contribution of CoInteract is twofold. First, we propose a \textbf{Human-Aware Mixture-of-Experts (MoE)} to improve hand and face quality. Using face and hand bounding boxes as supervision, a lightweight router learns to dispatch tokens to region-specialized experts that enhance structural fidelity with only a marginal increase in parameters.
Second, we introduce a \textbf{dual-stream co-generation paradigm} to enforce physical plausibility. An auxiliary HOI structure stream---where the human body is reduced to a silhouette while the object retains its RGB appearance---is jointly trained with the RGB stream within a shared DiT backbone. This forces the backbone to learn spatial and interaction relationships rather than relying on appearance cues alone. During training, the HOI stream attends to RGB tokens and its supervision regularizes the shared backbone; at inference, we keep only the RGB stream.
Extensive experiments demonstrate that CoInteract consistently outperforms existing methods in interaction plausibility, structural stability, and identity preservation.
Our contributions are summarized as follows:
\begin{itemize}
\item We propose \textbf{CoInteract}, a novel end-to-end framework for speech-driven HOI video synthesis that embeds human structural priors and interaction geometry constraints directly into the DiT backbone, ensuring both physical plausibility and structural consistency without relying on external preprocessing or post-processing.
\item We introduce a Human-Aware Mixture-of-Experts (MoE) that uses spatial supervision to guide lightweight specialized experts for hands and faces. This targeted processing ensures high structural fidelity and effectively reduces the  artifacts commonly seen in these critical regions, with minimal additional parameters.
\item We develop a Spatially-Structured Co-Generation paradigm using an asymmetric co-attention mask to embed physical interaction rules into the DiT. This approach forces the model to respect geometric constraints and substantially reduces hand-object interpenetration, while ensuring zero additional computational cost at inference.
\end{itemize}

\section{Related Works}

\subsection{Video Diffusion Models}
Video diffusion models~\cite{wan2025wan,kong2024hunyuanvideo,li2026skyreels,guo2023animatediff,blattmann2023stable} have evolved rapidly from image diffusion to temporally coherent video synthesis.
Early methods~\cite{guo2023animatediff,blattmann2023stable} extend image priors with temporal modules but often suffer from flickering and drift.
Subsequent works~\cite{voleti2024sv3d,wan2024grid} improve consistency via patch- or view-based designs.
Recent state-of-the-art approaches~\cite{wan2025wan,li2026skyreels} predominantly adopt DiT-style backbones~\cite{dit} to model spatiotemporal tokens with global attention.

Despite strong visual quality, RGB-centric video diffusion models remain fragile in complex human--object interaction (HOI) scenarios~\cite{liu2025hoigen}, where supervision provides weak constraints on contact geometry and body topology.
This often manifests as hand/face distortions and contact violations such as interpenetration.
To introduce additional structural cues, recent works explore multi-stream co-generation by jointly predicting auxiliary modalities (e.g., depth or flow) alongside RGB~\cite{chefer2025videojam,huang2025unityvideo}, or by explicitly injecting geometric constraints during training~\cite{wu2025geometry}.
However, these methods target general video synthesis and do not explicitly address HOI-specific challenges such as stable hand articulation under occlusion and physically plausible grasping.
Our work focuses on HOI and injects interaction-structure supervision and region-specific specialization into a shared DiT backbone.

\subsection{Audio-driven Human Animation}
Audio-driven human animation has achieved impressive results for talking heads and avatars, emphasizing lip synchronization and identity preservation~\cite{prajwal2020lip,zhou2020makelttalk,zhang2023sadtalker,wang2021audio2head,canonswap,pointtalk,yu2024gaussiantalker}.
Beyond faces, co-speech gesture generation maps speech to plausible body/hand motion, where diffusion models are effective in capturing one-to-many motion variability~\cite{yang2025infinitetalk,gao2025wans2v,lin2025omnihuman,gan2025omniavatar,guo2026dreamidomni}.
However, most prior work focuses on human-only motion and does not explicitly enforce hand--object contact constraints in rendered videos.

Recent studies recognize that hands and faces require dedicated treatment.
CyberHost~\cite{lin2025cyberhost} introduces region attention with learnable latent features to refine hand/face synthesis, and Make-Your-Anchor~\cite{huang2024makeyouranchor} applies post-hoc face enhancement.
These approaches typically act as external add-ons that are decoupled from the generative backbone.
In contrast, we embed a Human-Aware Mixture-of-Experts (MoE) directly into DiT blocks to specialize capacity for hands/faces \emph{within} the generation process, and further introduce HOI-structured supervision to improve physically plausible interactions.

\subsection{Human--Object Interaction Video Generation}
HOI video synthesis requires jointly modeling human motion, object manipulation, and physically plausible contact.
Existing approaches can be roughly grouped into two paradigms.

\textbf{(i) Multi-condition generation.}
These methods augment diffusion models with explicit pose and object-related structural controls~\cite{xu2026anchorcrafter,wang2025dreamactor,liu2025byteloom}.
For instance, AnchorCrafter~\cite{xu2026anchorcrafter} conditions on human pose and multi-view object features, while ByteLoom~\cite{liu2025byteloom} introduces geometric priors (e.g., relative coordinate maps) to improve spatial alignment.
While effective, such approaches rely on heavy preprocessing and external signals, and often do not internalize HOI constraints within the backbone.

\textbf{(ii) Multi-reference generation.}
Another line injects identity and product references without explicit geometry conditions.
This includes direct multi-reference video synthesis~\cite{chen2025humo,jiang2025vace,li2026skyreels,zhou2026omnishow} and two-stage pipelines that composite references via image editing~\cite{wu2025qwenimage,cai2025zimage} before speech-to-video generation~\cite{yang2025infinitetalk,gao2025wans2v,wang2025fantasytalking,gan2025omniavatar}.
Despite flexibility, these methods often lack HOI-specific structural supervision, leading to unstable hand articulation and implausible contact.

CoInteract addresses these limitations by formulating HOI synthesis as \emph{spatially-structured co-generation}:
it jointly trains an RGB stream with an auxiliary HOI structure stream to inject interaction geometry priors, and uses a Human-Aware MoE to stabilize structurally sensitive regions, while enabling zero-overhead inference by removing the auxiliary HOI branch.

\section{Method}
\begin{figure}[t]
    \includegraphics[width=\linewidth]{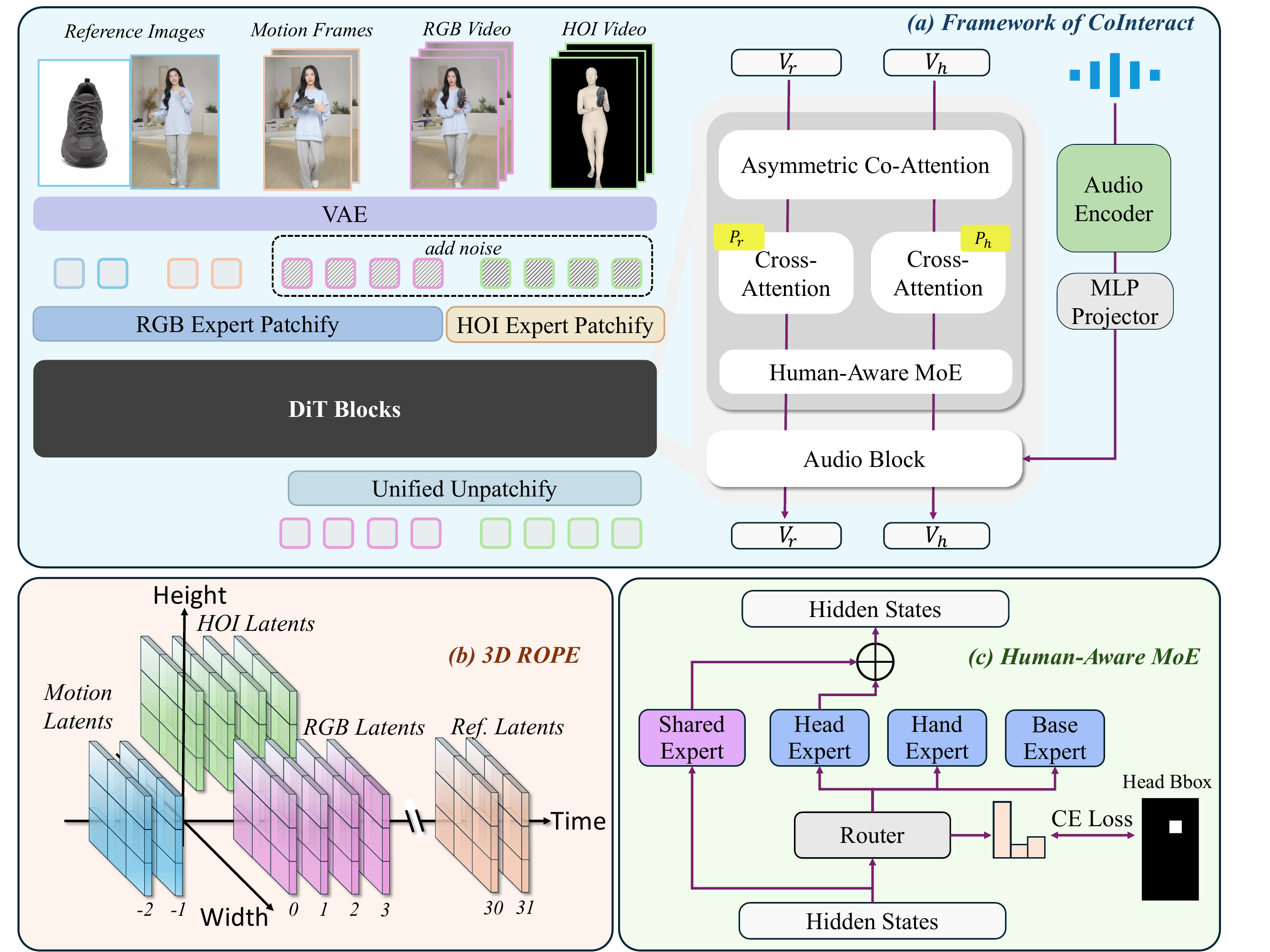}
    \caption{Overview of CoInteract.
    (a) The framework jointly generates RGB and HOI streams within a shared DiT backbone equipped with a Human-Aware MoE. An asymmetric co-attention mask enforces interaction-structure supervision during training; the HOI branch is removed at inference.
    $\mathbf{V}_{r,h}$ and $\mathbf{P}_{r,h}$ denote the hidden states and prompts of the RGB and HOI streams, respectively.
    (b) 3D RoPE assigns distinct spatiotemporal coordinates to motion, RGB, reference, and HOI latents.
    (c) The Human-Aware MoE employs a spatially supervised router to dispatch head and hand tokens to specialized experts.}
    \label{fig:pipeline}
\end{figure}

As illustrated in Fig.~\ref{fig:pipeline}, we present \textbf{CoInteract}, an end-to-end framework for speech-driven human--object interaction (HOI) video synthesis.
Given dual reference images (character identity $\mathcal{I}_{ref}$ and product $\mathcal{I}_{prod}$) together with motion frames $\mathcal{V}_{mot}$ that preserve temporal continuity~\cite{framepack,gao2025wans2v,luo2026filmweaver}, our goal is to synthesize HOI videos that are both structurally stable and physically plausible.
Unlike conventional video diffusion models that operate purely in RGB space, CoInteract explicitly injects interaction structure and body-level consistency into a shared Diffusion Transformer (DiT) backbone.

\subsection{Unified RGB--HOI Co-Generation}
Standard diffusion-based video models learn RGB-dominated spatiotemporal priors without explicit supervision on interaction geometry.
While effective for general synthesis, they provide weak constraints for contact topology and occlusion ordering, often producing HOI artifacts such as hand--object interpenetration and unstable hand articulation.

To address this, we introduce a unified co-generation paradigm (Fig.~\ref{fig:pipeline}(a)) in which an RGB appearance stream $\mathbf{z}_{r}$ and an auxiliary HOI structure stream $\mathbf{z}_{h}$ are jointly trained within a single DiT backbone.
The key insight is to train with a \emph{texture-stripped} HOI structure stream that preserves interaction geometry (mesh projection with fused object masks) while removing appearance cues, thereby guiding shared weights toward physically consistent representations.

\paragraph{HOI structure stream.}
We construct an auxiliary HOI structure stream as a silhouette-like 3-channel rendering obtained by projecting the recovered human mesh to the image plane and fusing the projected object mask.
This produces a pixel-aligned structural target that highlights interaction boundaries while discarding RGB texture.

\paragraph{Tokenization and shared backbone.}
The two streams are tokenized by \emph{modality-specific} patch embedding layers (with the same patch size) and then fed into shared DiT blocks.
The HOI stream uses a fixed descriptive prompt template to provide consistent semantic conditioning.
Within each DiT block, the two streams share all transformer parameters but employ stream-specific modulation parameters (scale and shift in adaptive layer normalization), enabling a single backbone to specialize feature statistics for appearance versus structure without duplicating the full model.

\paragraph{Joint flow-matching objective.}
We optimize the model with a joint flow-matching objective supervising both streams:
\begin{equation}
\mathcal{L}_{flow} = \mathcal{L}_{r} + \lambda_h \mathcal{L}_{h}.
\end{equation}

\begin{equation}
\begin{aligned}
\mathcal{L}_{r} &= \mathbb{E}_{t,\mathbf{z}_0,\mathbf{z}_1}\!\left[\left\|\mathbf{v}_r - \mathbf{v}_\theta(\mathbf{z}_{r,t}, t, \mathbf{c})\right\|_2^2\right], \\
\mathcal{L}_{h} &= \mathbb{E}_{t,\mathbf{z}_0,\mathbf{z}_1}\!\left[\left\|\mathbf{v}_h - \mathbf{v}_\theta(\mathbf{z}_{h,t}, t, \mathbf{c})\right\|_2^2\right].
\end{aligned}
\end{equation}

where $\mathbf{v}$ denotes the target velocity field, $t$ is the diffusion timestep, and $\mathbf{c}$ denotes conditioning (text, audio, dual reference images, and motion latents).
We set $\lambda_h = 1$ unless otherwise stated.

\subsubsection{Multi-Modal Coordinate Assignment via 3D RoPE}
To seamlessly integrate heterogeneous modalities---ranging from historical motion and static references to dual-stream generative latents---we assign each token a 3D coordinate $(h,w,t)$ encoded by 3D Rotary Positional Encoding (3D RoPE).
As illustrated in Fig.~\ref{fig:pipeline}(b), our coordinate assignment enforces the following inductive biases:
\begin{itemize}
    \item \textbf{Spatial coordinates for dual streams.}
    To preserve pixel-level correspondence between RGB and HOI, we concatenate the two streams along the width dimension, assigning distinct horizontal coordinates---\eg $w\in[0,W]$ for RGB and $w\in[-W,0]$ for HOI---while sharing identical height and time indices.
    This allows the model to learn cross-stream alignment through relative positional distances.

    \item \textbf{Temporal causality and reference anchoring.}
    We explicitly structure the temporal axis to distinguish between motion context, generated video, and identity reference:
    \begin{enumerate}
        \item \textit{Historical motion ($t<0$).}
        Past motion frames are assigned negative temporal indices (e.g., $t\in\{-N,\dots,-1\}$), placing them logically before the current generation window to encourage causal motion continuity.
        \item \textit{Reference anchoring ($t\gg T$).}
        Static reference images are mapped to a far-field temporal location (e.g., $t=30,31$) with a significant offset, encouraging the model to treat them as global identity anchors rather than adjacent frames.
    \end{enumerate}
\end{itemize}
Formally, the position of any token $x$ is encoded as:
\begin{equation}
\text{Pos}(x_{i,j,k}) = \text{RoPE}_{3D}(h_i, w'_j, t_k),
\end{equation}
where $w'_j$ accounts for the virtual width shift in the HOI stream.
This unified mapping encourages the attention mechanism to respect structural and temporal relationships across all inputs.

\subsubsection{Two-Stage Asymmetric Co-Attention}
\begin{figure}[t]
    \centering
    \includegraphics[width=\linewidth]{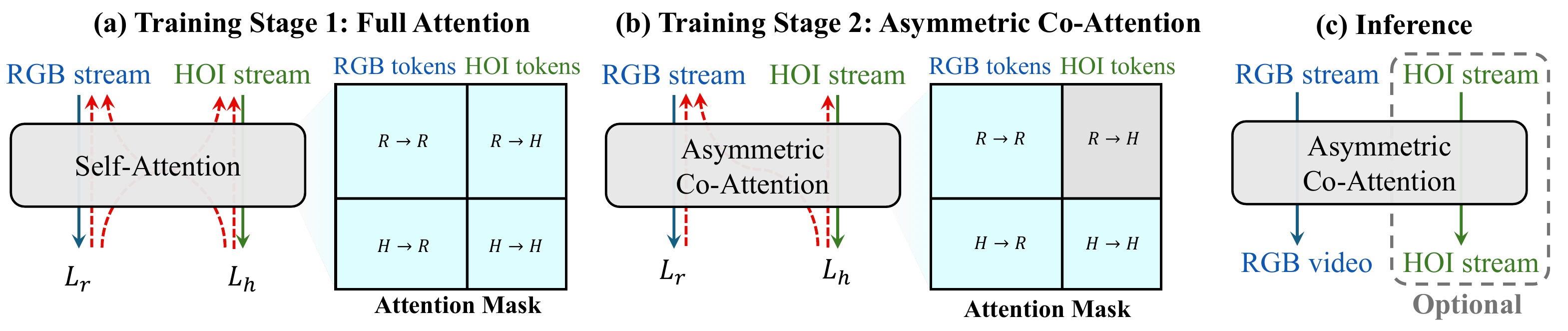}
    \caption{Two-stage training and inference (DiT blocks omitted for clarity).
    Stage~1 uses full attention to establish coupling between the RGB and HOI streams.
    Stage~2 applies an asymmetric co-attention mask, enabling removal of the HOI branch at inference for efficiency.
    $\mathcal{L}_{r}$ and $\mathcal{L}_{h}$ denote the flow-matching losses for the RGB and HOI streams, respectively, and red dashed arrows indicate gradient flow into shared DiT parameters.}
    \label{fig:co-attention}
\end{figure}

To inject interaction-structure supervision while maintaining inference efficiency, we adopt a two-stage training strategy with an \textbf{Asymmetric Co-Attention} mechanism (Fig.~\ref{fig:co-attention}).
In Stage~1, standard bidirectional attention is applied across both streams for rapid convergence, allowing the model to learn global dependencies between appearance and interaction structure.
In Stage~2, we enforce an asymmetric attention mask.
Let $\mathcal{T}_r$ and $\mathcal{T}_h$ denote the token sets for the RGB and HOI streams, respectively, and let rows correspond to queries and columns to keys.
The mask $\mathbf{M}$ is defined as:
\begin{equation}
\mathbf{M}_{i,j} =
\begin{cases}
1, & \text{if } i \in \mathcal{T}_r,\; j \in \mathcal{T}_r,\\
1, & \text{if } i \in \mathcal{T}_h,\; j \in \mathcal{T}_r \cup \mathcal{T}_h,\\
0, & \text{otherwise.}
\end{cases}
\end{equation}
Under this mask, RGB queries attend only to RGB tokens, making the RGB pathway independent of the HOI branch and thus removable at inference with \emph{zero} overhead.
HOI queries, conversely, attend to both streams, leveraging cleaner RGB features to predict interaction structure.
Crucially, $\mathcal{L}_{h}$ backpropagates through the HOI$\leftarrow$RGB cross-attention into the \emph{shared} DiT parameters; since the RGB stream reuses the same parameters at inference, this transfers interaction-structure supervision to the RGB generator even when the HOI branch is removed.

\subsection{Human-Aware Mixture-of-Experts}
While structured co-generation injects interaction priors into the backbone, hands and faces may still exhibit artifacts due to their high-frequency detail and articulation complexity.
We therefore incorporate a \textbf{Human-Aware Mixture-of-Experts (MoE)} module~\cite{shi2025diffmoe,ditmoe} that routes tokens to region-specialized experts via a spatially supervised router $\mathcal{R}$ (Fig.~\ref{fig:pipeline}(c)).
We include a \emph{Shared} expert that reuses the original DiT FFN as a shortcut path, and three lightweight experts (\emph{Head}, \emph{Hand}, \emph{Base}) implemented as small FFNs, introducing a modest parameter overhead.
This enables dedicated capacity for anatomically sensitive regions, improving hand sharpness and face identity consistency without degrading general synthesis quality.

\paragraph{Spatially supervised routing.}
To prevent router optimization from interfering with the DiT representation learning, we apply a stop-gradient operation $sg[\cdot]$ to hidden states before routing.
The routing probability for token $x_i$ is computed as:
\begin{equation}
G(x_i) = \text{Softmax}(\mathcal{W}_g \cdot sg[\mathbf{h}_i]).
\end{equation}
Using face and hand bounding boxes, the router assigns tokens inside the corresponding regions to $\mathcal{E}_{head}$ or $\mathcal{E}_{hand}$, while remaining tokens are processed by the base expert.
We enforce specialization via a cross-entropy routing loss:
\begin{equation}
\mathcal{L}_{route} = - \sum_{i} \sum_{k \in \{head, hand, base\}} \mathbb{1}(y_i = k) \log(G(x_i)_k),
\end{equation}
where $y_i$ is the ground-truth region label and $\mathbb{1}(\cdot)$ is the indicator function.
The total training objective is:
\begin{equation}
\mathcal{L}_{total} = \mathcal{L}_{flow} + \eta \mathcal{L}_{route}.
\end{equation}

\subsection{Data Curation and Representation}
\label{sec:curation}
\begin{figure}[t]
    \centering
    \includegraphics[width=\linewidth]{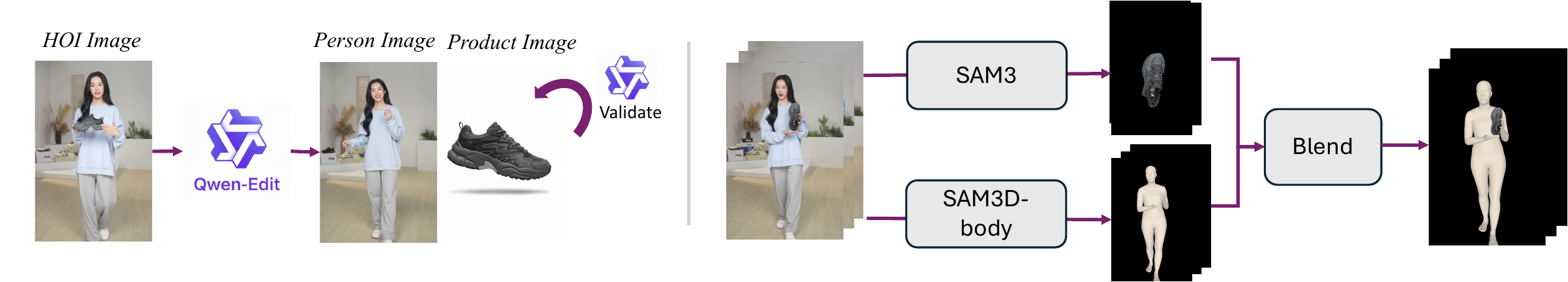}
    \caption{Preprocessing pipeline for constructing paired RGB and HOI-structure training data.}
    \label{fig:curation}
\end{figure}

To facilitate structure-aware training, we transform raw HOI videos into paired RGB and HOI-structure representations (Fig.~\ref{fig:curation}).
We first decouple entities using Qwen-Edit~\cite{wu2025qwenimage} to create independent person and product references, followed by a validation module that filters mismatched (source image, person, object) triplets.
For geometric supervision, we use SAM3~\cite{carion2025sam3} to obtain object masks and SAM3D-body~\cite{sambody} to recover a human mesh, which is then projected to the image plane.
We fuse the projected human rendering with the object mask to form a texture-stripped HOI structure stream $\mathbf{V}_h$.
Since this HOI stream is geometry-focused, the model is encouraged to learn spatial and \emph{interaction} relationships rather than exploiting appearance shortcuts.
Finally, both RGB video $\mathbf{V}_r$ and HOI stream $\mathbf{V}_h$ are encoded into a shared latent space via a pre-trained VAE for dual-stream training.
Additionally, we use off-the-shelf detectors~\cite{mediapipe,shan2020understanding} to obtain face and hand bounding boxes, which provide explicit supervision for the MoE router during training.

\section{Experiments}

\subsection{Training Details}

\noindent\textbf{Dataset.}
We curate a large-scale HOI video dataset following Section~\ref{sec:curation}, comprising 40 hours of product demonstration and live-streaming videos. After quality filtering, 12K high-quality clips are retained with paired RGB--HOI representations, hand/face bounding boxes, and silhouette masks. 
A held-out test set of 50 clips covers diverse product categories and unseen identities.

\noindent\textbf{Implementation Details.}
CoInteract is initialized from WanS2V~\cite{gao2025wans2v}. The Human-Aware MoE consists of four experts: one Shared expert that reuses the original DiT FFN, and three lightweight experts (Head, Hand, Base) each implemented as a light FFN with hidden dimension 256. The router is a two-layer MLP trained with spatial supervision. AdamW~\cite{kingma2014adam} is used with learning rate $1\times10^{-4}$ and cosine annealing. Training proceeds in two stages: full bidirectional self-attention with 5K iterations, followed by asymmetric co-attention with 2K iterations. Loss weights are $\lambda_{h}=1$ and $\eta=1$. The HOI branch is discarded at inference, incurring zero additional cost.  As for inference settings, we set CFG with 5, inference steps with 40, and generate videos at 480p resolution.

\subsection{Quantitative Comparison}

\noindent\textbf{Baselines.}
We compare CoInteract against AnchorCrafter~\cite{xu2026anchorcrafter}, Phantom~\cite{liu2025phantom}, Humo~\cite{chen2025humo}, InteractAvatar~\cite{zhang2026interact}, SkyReels-V3~\cite{li2026skyreels}, and VACE~\cite{jiang2025vace}. All methods use identical reference images and audio inputs. Since InteractAvatar does not natively support separate injection of identity and product references, we first use Qwen-Image~\cite{wu2025qwenimage} to composite both references into a single image before feeding it to the model; its results therefore partly benefit from the image editing model's compositing capability. For AnchorCrafter, we follow the authors' preprocessing pipeline to prepare per-frame pose and object annotations for all test samples.

\noindent\textbf{Evaluation Metrics.}
We evaluate from four complementary aspects.

\textit{Video Quality.}
\textit{AES}$\uparrow$ measures perceptual aesthetics via the LAION aesthetic predictor~\cite{schuhmann2022laion}.
\textit{IQ}$\uparrow$ uses MUSIQ~\cite{ke2021musiq} for frame-level perceptual quality.
\textit{Smooth}$\uparrow$ computes CLIP cosine similarity between consecutive frames.
All three metrics are computed via VBench~\cite{huang2024vbench}.

\textit{Human--Object Interaction.}
Standardized benchmarks for evaluating HOI video quality remain scarce; we therefore rely primarily on two automatic metrics alongside a perceptual user study (Sec.~\ref{sec:user_study}).
\textit{VLM-QA}$\uparrow$ employs Gemini-3-Pro~\cite{gemini2024}, one of the most capable video perception models, to assess HOI plausibility. We design a structured questionnaire of 50 binary (0/1) questions that probe interaction rationality; the per-video score is the fraction of positive responses, averaged over the test set.
\textit{HQ}$\uparrow$ (Hand Quality) computes the mean confidence of hand keypoints detected by DWPose~\cite{yang2023dwpose}, averaged over all frames in the generated video. Higher scores indicate more structurally plausible and clearly rendered hands.

\textit{Reference Consistency.}
\textit{DINO}$_\textit{id}$$\uparrow$ measures DINOv2~\cite{oquab2024dinov2} cosine similarity between the identity reference $\mathcal{I}_{ref}$ and generated character crops.
\textit{DINO}$_\textit{obj}$$\uparrow$ measures DINOv2 similarity between the product reference $\mathcal{I}_{prod}$ and generated object regions.
\textit{FaceSim}$\uparrow$ computes ArcFace~\cite{deng2019arcface} cosine similarity between reference face embeddings and generated face embeddings averaged across frames.

\textit{Audio-Visual Alignment.}
\textit{Sync}$_\textit{conf}$$\uparrow$ is the lip-sync confidence score~\cite{chung2016syncnet}.

\begin{figure}[t]
    \centering
    \includegraphics[width=\linewidth]{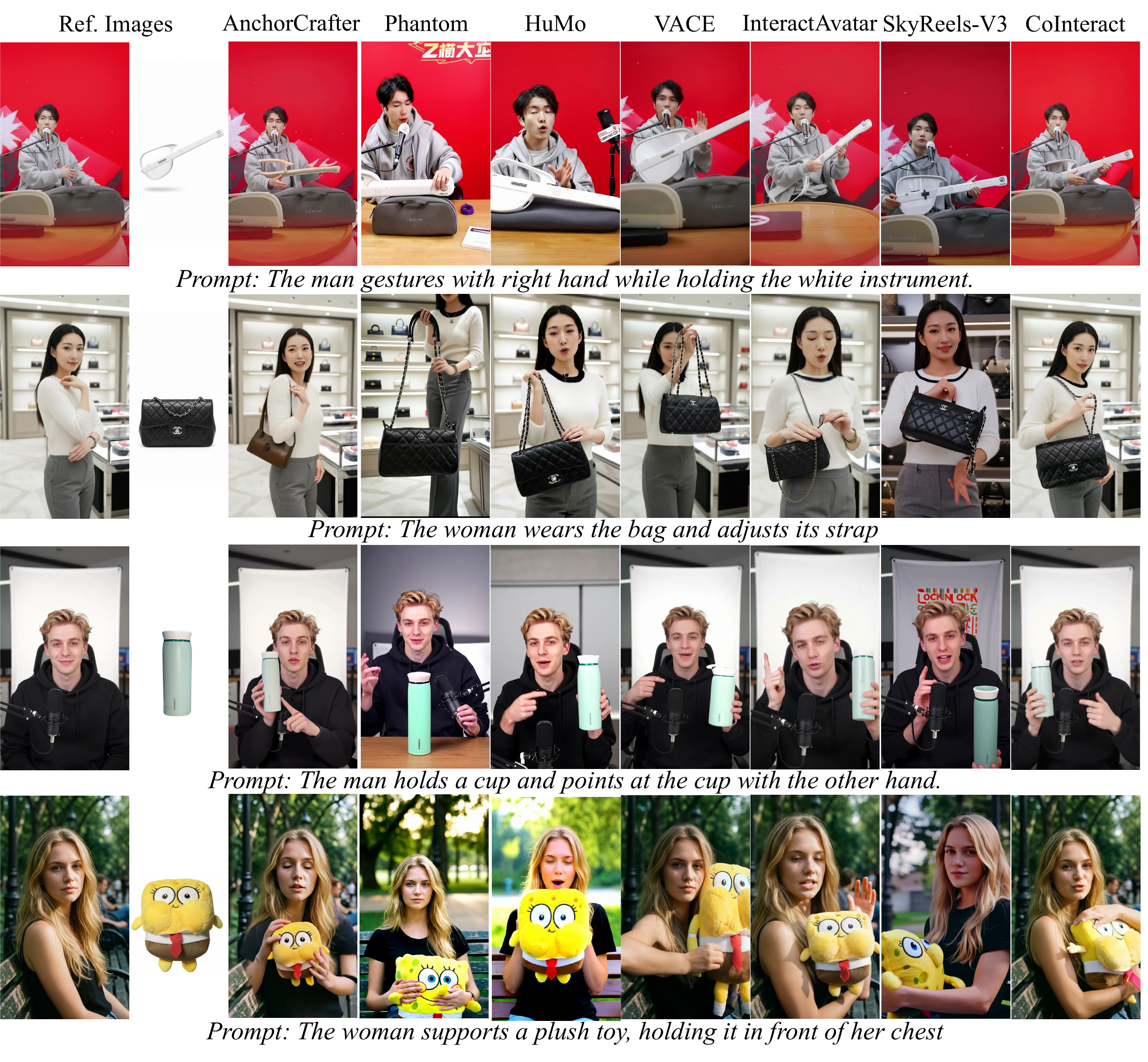}
    \caption{Qualitative comparison with existing methods, CoInteract preserves higher interaction fidelity and better adheres to the input prompts. (Zoom in for details.)}
    \label{fig:qual}
\end{figure}

\noindent\textbf{Results.}
Table~\ref{tab:main} presents quantitative comparisons. CoInteract achieves the best or competitive results across most metrics. In particular, it obtains the highest VLM-QA (0.72) and HQ (0.724), confirming superior interaction plausibility and hand structural stability. It also leads in identity preservation (DINO$_\text{id}$: 0.671, FaceSim: 0.696) and temporal coherence (Smooth: 0.9951). Phantom and Humo score slightly higher on AES, partly because they tend to hallucinate novel backgrounds that happen to look aesthetically pleasing but deviate from the input reference (see Fig.~\ref{fig:qual}); CoInteract instead faithfully preserves the reference scene, which trades marginal aesthetic scores for stronger consistency.

\begin{table*}[t]
    \centering
        \caption{ 
        Quantitative comparison on the HOI video generation test set.
        \textbf{Bold} indicates the best result; \underline{underline} indicates the second best; ``---'' indicates unsupported.
        }
        \label{tab:main}
        \resizebox{\textwidth}{!}{%
        \begin{tabular}{l ccc cc ccc c}
        \toprule
        \multirow{2}{*}{\textbf{Method}}
        & \multicolumn{3}{c}{\textit{Video Quality}}
        & \multicolumn{2}{c}{\textit{HOI}}
        & \multicolumn{3}{c}{\textit{Reference}}
        & \textit{Audio} \\
        \cmidrule(lr){2-4}
        \cmidrule(lr){5-6}
        \cmidrule(lr){7-9}
        \cmidrule(lr){10-10}
        & AES & IQ & Smooth
        & VLM-QA & HQ
        & DINO$_\text{id}$ & DINO$_\text{obj}$ & FaceSim
        & Sync$_\text{conf}$ \\
        \midrule
        AnchorCrafter~\cite{xu2026anchorcrafter} & 0.448 & 0.643 & 0.9743 & 0.22 & 0.596 & 0.538 & 0.453 & 0.487 & --- \\
        Phantom~\cite{liu2025phantom} & \textbf{0.579} & 0.724 & 0.9916 & 0.50 & 0.650 & 0.654 & 0.595 & 0.593 & --- \\
        Humo~\cite{chen2025humo} & \underline{0.565} & \underline{0.741} & 0.9919 & 0.56 & 0.664 & 0.643 & \underline{0.629} & 0.618 & 5.71 \\
        VACE~\cite{jiang2025vace} & 0.530 & 0.733 & 0.9904 & 0.46 & 0.627 & 0.623 & \textbf{0.635} & 0.647 & --- \\
        InteractAvatar~\cite{zhang2026interact} & 0.528 & 0.722 & \underline{0.9938} & \underline{0.62} & \underline{0.696} & \underline{0.658} & 0.608 & \underline{0.681} & 5.82 \\
        SkyReels-V3~\cite{li2026skyreels} & 0.563 & 0.720 & 0.9861 & 0.44 & 0.626 & 0.637 & 0.564 & 0.569 & --- \\
        \midrule
        \textbf{CoInteract}
            & 0.554 & \textbf{0.749} & \textbf{0.9951}
            & \textbf{0.72} & \textbf{0.724}
            & \textbf{0.671} & 0.624 & \textbf{0.696}
            & \textbf{5.87} \\
        \bottomrule
    \end{tabular}}
\end{table*}

\subsection{Qualitative Results}

Fig.~\ref{fig:qual} presents qualitative comparisons across diverse scenarios. CoInteract consistently produces videos with coherent hand articulation, natural product grasping, and faithful prompt adherence. Other baselines exhibit varying degrees of deficiency in HOI plausibility, prompt compliance, and background consistency---common failure modes include hand-object interpenetration, inconsistent product appearance, and background deviation from the reference. AnchorCrafter performs noticeably better on the last two cases, which correspond to objects in its training set; on the first two unseen-object cases, it suffers from identity drift and unnatural interaction boundaries, revealing limited generalization. InteractAvatar benefits from Qwen-Image compositing, which provides a strong initial frame with correct object placement; however, as generation progresses, it still produces HOI plausibility issues such as unnatural grasping poses (e.g., last row). In contrast, CoInteract maintains physically plausible interaction and structural stability throughout the full sequence.

\subsubsection{Visualization of HOI branch and Human-Aware MoE}
\begin{figure}[t]
    \centering
    \includegraphics[width=0.85\linewidth]{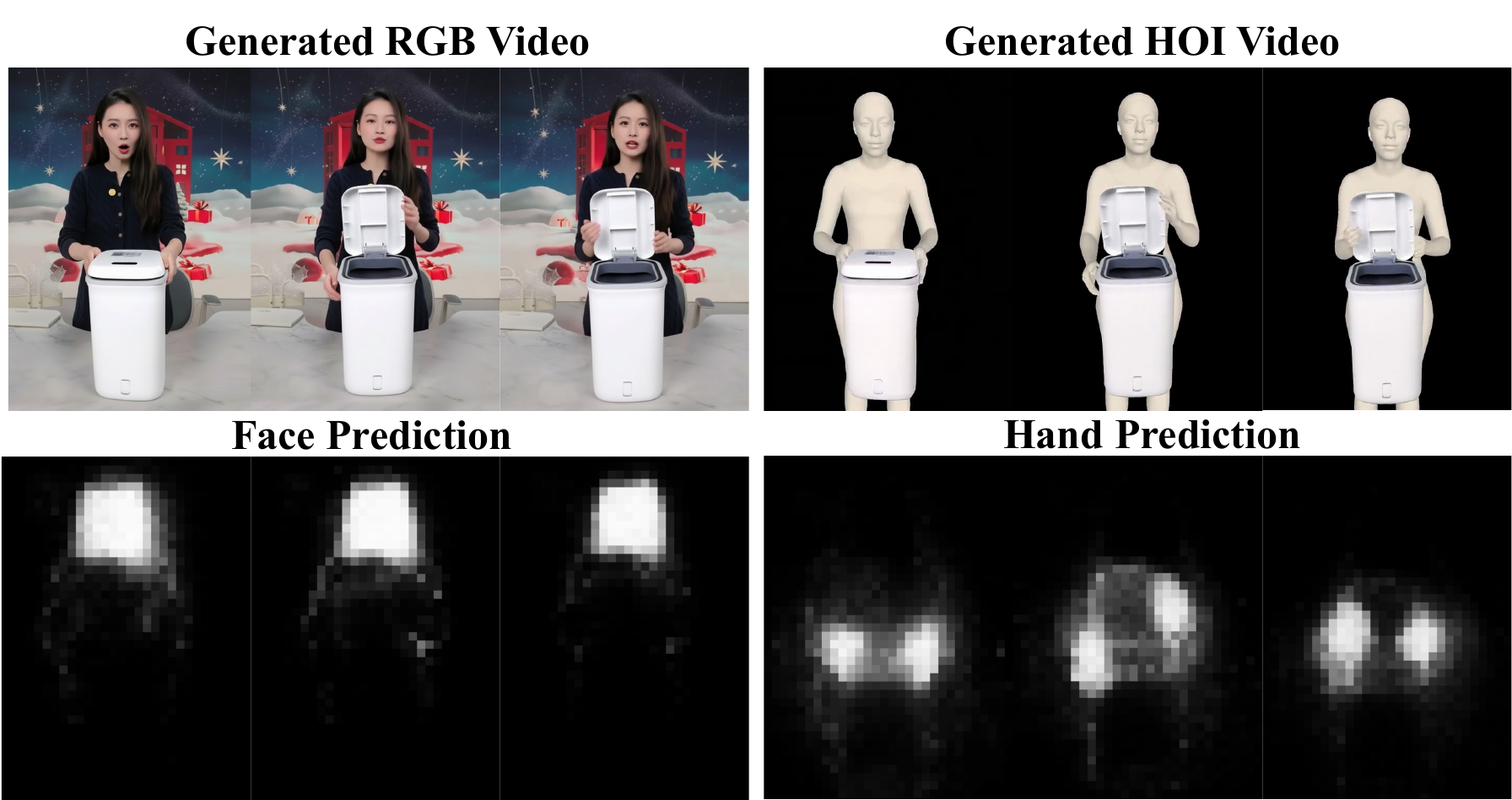}
    \caption{Visualization of Dual-Stream Co-generation and MoE Routing. The top rows display the generated RGB video and the corresponding auxiliary HOI representation, demonstrating precise spatio-temporal alignment and structural consistency during complex interactions (e.g., opening a bin). The bottom rows show the routing heatmaps for face and hand experts from the MoE router.}
    \label{fig:qualitative2}
\end{figure}

To further investigate CoInteract's internal mechanisms, we visualize the dual-stream generation and MoE routing behavior in Fig.~\ref{fig:qualitative2}. The HOI stream maintains precise spatio-temporal synchronization with the RGB stream, providing a consistent geometric scaffold that mitigates hand-object interpenetration even during drastic motions such as opening a bin lid. The routing heatmaps confirm that the router accurately isolates face and hand tokens and dispatches them to specialized experts, maintaining high-frequency structural fidelity even under rapid movement.
\subsection{User Study}
\label{sec:user_study}

We conducted a perceptual user study with 24 evaluators recruited via crowd-sourcing.
Each evaluator was shown 10 randomly sampled test cases. For each case, all methods were
provided with identical inputs, and the resulting videos were presented in a blind randomized order.
Evaluators were asked to rank the methods (lower is better) according to three criteria:
Object Consistency, Human/Background Consistency, and Interaction Plausibility.
\begin{table}[t]
\centering
\caption{User study (mean rank, lower is better).}
\label{tab:user}
\resizebox{\linewidth}{!}{%
\begin{tabular}{l ccccccc}
\toprule
\textbf{Criterion}
    & AnchorCrafter & Phantom & Humo & VACE & InteractAvatar & SkyReels-V3 & \textbf{CoInteract} \\
\midrule
Obj. Consist.$\downarrow$  & 6.08 & 4.13 & 4.42 & 3.54 & 3.08 & 4.58 & \textbf{2.17} \\
Hum/BG$\downarrow$         & 6.28 & 4.38 & 4.21 & 3.46 & 2.92 & 4.83 & \textbf{1.92} \\
Interact.$\downarrow$      & 6.55 & 4.29 & 3.92 & 3.58 & 3.33 & 4.54 & \textbf{1.79} \\
\bottomrule
\end{tabular}}
\end{table}
Table~\ref{tab:user} reports the mean rank. CoInteract achieves the best (lowest) mean rank
across all criteria, with the largest advantage on Interaction Plausibility, consistent with
our HOI-focused design.
\subsection{Ablation Study}
\begin{table}[t]
    \centering
    \caption{
        Ablation study on core components.
    }
    \label{tab:ablation}
    \resizebox{\linewidth}{!}{%
    \begin{tabular}{l ccc cc ccc c c}
        \toprule
        \multirow{2}{*}{\textbf{Variant}}
            & \multicolumn{3}{c}{\textit{Video Quality}}
            & \multicolumn{2}{c}{\textit{HOI}}
            & \multicolumn{3}{c}{\textit{Reference}}
            & \textit{Audio}
            & \\
        \cmidrule(lr){2-4}
        \cmidrule(lr){5-6}
        \cmidrule(lr){7-9}
        \cmidrule(lr){10-10}
        \cmidrule(lr){11-11}
            & AES & IQ & Smooth
            & VLM-QA & HQ
            & DINO$_\text{id}$ & DINO$_\text{obj}$ & FaceSim
            & Sync$_\text{conf}$
            & Infer.\ Cost \\
        \midrule
        w/o MoE
            & 0.541 & 0.736 & 0.993
            & 0.66 & 0.658
            & 0.659 & 0.611 & 0.662
            & 5.64 & $1.00\times$ \\
        w/o Co-Gen
            & 0.536 & 0.753 & 0.991
            & 0.48 & 0.706
            & 0.664 & 0.597 & 0.678
            & 5.86 & $1.04\times$ \\
        w/o Asym.\ Mask
            & 0.548 & 0.742 & 0.994
            & 0.76 & 0.738
            & 0.668 & 0.618 & 0.689
            & 5.81 & $4.13\times$ \\
        \midrule
        Full Model
            & 0.554 & 0.749 & 0.995
            & 0.72 & 0.724
            & 0.671 & 0.624 & 0.696
            & 5.87 & $1.04\times$ \\
        \bottomrule
    \end{tabular}}
\end{table}
\begin{figure}[t]
    \centering
    \includegraphics[width=\linewidth]{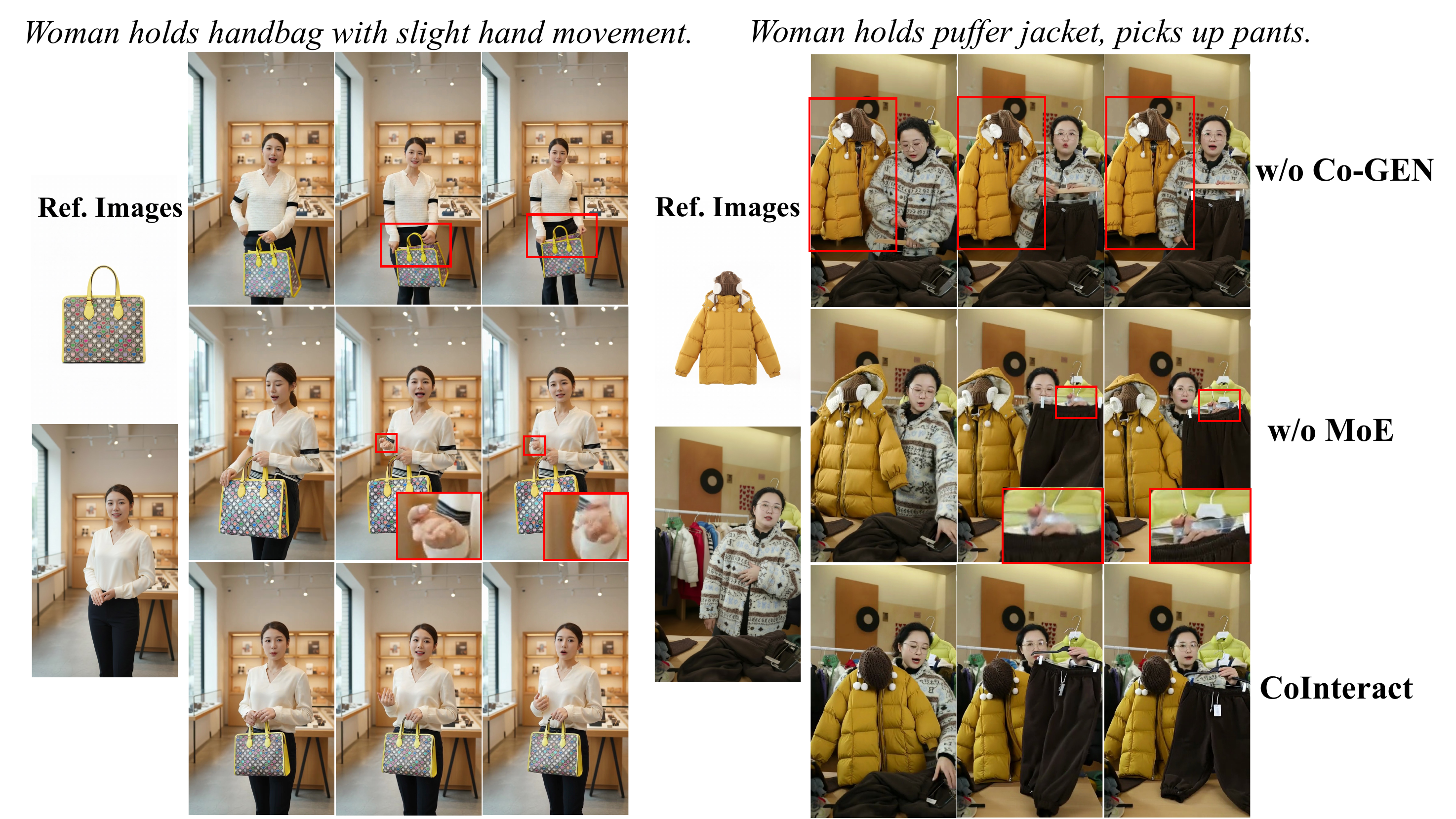}
    \caption{Qualitative comparison of ablation variants. The absence of unified HOI co-generation yields interactions that lack physical plausibility. Conversely, removing the Human-Aware MoE induces structural collapse and artifacts in high-frequency regions like hands. (Zoom in for details.)}
    \label{fig:ablation}
\end{figure}
We systematically ablate each core component via four model variants:
(1) w/o MoE, replacing the Human-Aware MoE with a standard FFN to disable expert specialization;
(2) w/o Co-Gen, removing the HOI stream entirely to reduce the model to a single-stream RGB baseline without structural supervision;
(3) w/o Asym.\ Mask, replacing the Stage-2 asymmetric co-attention mask with standard bidirectional self-attention, requiring the HOI branch to be retained at inference; and
(4) Full Model, the complete CoInteract.
All variants share identical training configurations.

Quantitative results are reported in Table~\ref{tab:ablation}. Removing MoE (row~1) notably degrades HQ ($0.724\!\to\!0.658$) and FaceSim ($0.696\!\to\!0.662$), confirming its role in fine-grained structural fidelity. Removing the HOI stream (row~2) causes the largest drop in VLM-QA ($0.72\!\to\!0.48$, $-$33.3\%), demonstrating that the auxiliary stream is essential for internalizing physical interaction constraints. Retaining the HOI branch at inference (row~3) slightly surpasses the full model on VLM-QA (0.76) and HQ (0.738), as expected from direct structural guidance; however, it inflates inference cost to $4.13\times$ due to the doubled token count. The MoE introduces only a $1.04\times$ overhead relative to the MoE-free baseline, confirming its minimal impact on inference efficiency. Our asymmetric strategy trades marginal interaction gains for a dramatic efficiency improvement at near-zero additional inference cost. Qualitative results in Fig.~\ref{fig:ablation} corroborate these findings: removing co-generation yields physically implausible interactions, while removing MoE leads to hand collapse and blurred facial details. 

\section{Conclusion}
This paper presents \textbf{CoInteract}, a structure-aware framework for speech-driven human-object interaction (HOI) video synthesis that prioritizes structural integrity and physical consistency. The framework introduces a Human-Aware Mixture-of-Experts (MoE) to enhance the fidelity of hands and faces through a spatially-supervised routing policy. Furthermore, the Spatially-Structured Co-Generation paradigm, utilizing an asymmetric co-attention mask, allows the model to learn physical interaction priors during training. This architecture effectively reduces hand-object interpenetration and geometric misalignment while maintaining a zero-overhead inference path. Extensive experiments demonstrate that CoInteract consistently outperforms existing methods in interaction plausibility and structural stability, advancing the quality of HOI video generation.



%
%
\bibliographystyle{splncs04}
\bibliography{main}

@String(CVPR  = {IEEE Conf. Comput. Vis. Pattern Recog.})

@String(AAAI  = {AAAI})

@String(TOG   = {ACM Trans. Graph.})

@String(CVPR  = {CVPR})

@String(TOG   = {ACM TOG})

@inproceedings{pointtalk,
  title={Pointtalk: Audio-driven dynamic lip point cloud for 3d gaussian-based talking head synthesis},
  author={Xie, Yifan and Feng, Tao and Zhang, Xin and Luo, Xiangyang and Guo, Zixuan and Yu, Weijiang and Chang, Heng and Ma, Fei and Yu, Fei Richard},
  booktitle={Proceedings of the AAAI Conference on Artificial Intelligence},
  volume={39},
  number={8},
  pages={8753--8761},
  year={2025}
}

@inproceedings{canonswap,
  title={Canonswap: High-fidelity and consistent video face swapping via canonical space modulation},
  author={Luo, Xiangyang and Zhu, Ye and Liu, Yunfei and Lin, Lijian and Wan, Cong and Cai, Zijian and Li, Yu and Huang, Shao-Lun},
  booktitle={Proceedings of the IEEE/CVF International Conference on Computer Vision},
  pages={10064--10074},
  year={2025}
}

@article{xue2025human,
  title={Human motion video generation: A survey},
  author={Xue, Haiwei and Luo, Xiangyang and Hu, Zhanghao and Zhang, Xin and Xiang, Xunzhi and Dai, Yuqin and Liu, Jianzhuang and Zhang, Zhensong and Li, Minglei and Yang, Jian and others},
  journal={IEEE Transactions on Pattern Analysis and Machine Intelligence},
  year={2025},
  publisher={IEEE}
}

@inproceedings{luo2026filmweaver,
  title={Filmweaver: Weaving consistent multi-shot videos with cache-guided autoregressive diffusion},
  author={Luo, Xiangyang and Li, Qingyu and Liu, Xiaokun and Qin, Wenyu and Yang, Miao and Wang, Meng and Wan, Pengfei and Zhang, Di and Gai, Kun and Huang, Shao-Lun},
  booktitle={Proceedings of the AAAI Conference on Artificial Intelligence},
  volume={40},
  number={9},
  pages={7689--7697},
  year={2026}
}

@article{wan2024grid,
  title={Grid: Omni visual generation},
  author={Wan, Cong and Luo, Xiangyang and Luo, Hao and Cai, Zijian and Song, Yiren and Zhao, Yunlong and Bai, Yifan and Wang, Fan and He, Yuhang and Gong, Yihong},
  journal={arXiv preprint arXiv:2412.10718},
  year={2024}
}

@article{chen2025humo,
  title={Humo: Human-centric video generation via collaborative multi-modal conditioning},
  author={Chen, Liyang and Ma, Tianxiang and Liu, Jiawei and Li, Bingchuan and Chen, Zhuowei and Liu, Lijie and He, Xu and Li, Gen and He, Qian and Wu, Zhiyong},
  journal={arXiv preprint arXiv:2509.08519},
  year={2025}
}

@inproceedings{liu2025phantom,
  title={Phantom: Subject-consistent video generation via cross-modal alignment},
  author={Liu, Lijie and Ma, Tianxiang and Li, Bingchuan and Chen, Zhuowei and Liu, Jiawei and Li, Gen and Zhou, Siyu and He, Qian and Wu, Xinglong},
  booktitle={Proceedings of the IEEE/CVF International Conference on Computer Vision},
  pages={14951--14961},
  year={2025}
}

@article{carion2025sam3,
  title={Sam 3: Segment anything with concepts},
  author={Carion, Nicolas and Gustafson, Laura and Hu, Yuan-Ting and Debnath, Shoubhik and Hu, Ronghang and Suris, Didac and Ryali, Chaitanya and Alwala, Kalyan Vasudev and Khedr, Haitham and Huang, Andrew and others},
  journal={arXiv preprint arXiv:2511.16719},
  year={2025}
}

@article{kingma2014adam,
  title={Adam: A method for stochastic optimization},
  author={Kingma, Diederik P and Ba, Jimmy},
  journal={arXiv preprint arXiv:1412.6980},
  year={2014}
}

@article{wan2025wan,
  title={Wan: Open and advanced large-scale video generative models},
  author={Wan, Team and Wang, Ang and Ai, Baole and Wen, Bin and Mao, Chaojie and Xie, Chen-Wei and Chen, Di and Yu, Feiwu and Zhao, Haiming and Yang, Jianxiao and others},
  journal={arXiv preprint arXiv:2503.20314},
  year={2025}
}

@inproceedings{jiang2025vace,
  title={Vace: All-in-one video creation and editing},
  author={Jiang, Zeyinzi and Han, Zhen and Mao, Chaojie and Zhang, Jingfeng and Pan, Yulin and Liu, Yu},
  booktitle={Proceedings of the IEEE/CVF International Conference on Computer Vision},
  pages={17191--17202},
  year={2025}
}

@article{xu2026anchorcrafter,
  title={AnchorCrafter: Animate Cyber-Anchors Selling Your Products via Human-Object Interacting Video Generation},
  author={Xu, Ziyi and Huang, Ziyao and Cao, Juan and Zhang, Yong and Cun, Xiaodong and Shuai, Qing and Wang, Yuchen and Bao, Linchao and Tang, Fan},
  journal={IEEE Transactions on Visualization and Computer Graphics},
  year={2026},
  publisher={IEEE}
}

@article{wang2025dreamactor,
  title={Dreamactor-h1: High-fidelity human-product demonstration video generation via motion-designed diffusion transformers},
  author={Wang, Lizhen and Xia, Zhurong and Hu, Tianshu and Wang, Pengrui and Wei, Pengfei and Zheng, Zerong and Zhou, Ming and Zhang, Yuan and Gao, Mingyuan},
  journal={arXiv preprint arXiv:2506.10568},
  year={2025}
}

@article{liu2025byteloom,
  title={ByteLoom: Weaving Geometry-Consistent Human-Object Interactions through Progressive Curriculum Learning},
  author={Liu, Bangya and Gong, Xinyu and Zhao, Zelin and Song, Ziyang and Lu, Yulei and Wu, Suhui and Zhang, Jun and Banerjee, Suman and Zhang, Hao},
  journal={arXiv preprint arXiv:2512.22854},
  year={2025}
}

@article{zhang2026interact,
  title={Making Avatars Interact: Towards Text-Driven Human-Object Interaction for Controllable Talking Avatars},
  author={Zhang, Youliang and Zhou, Zhengguang and Yu, Zhentao and Huang, Ziyao and Hu, Teng and Liang, Sen and Zhang, Guozhen and Peng, Ziqiao and Li, Shunkai and Chen, Yi and others},
  journal={arXiv preprint arXiv:2602.01538},
  year={2026}
}

@article{huang2025unityvideo,
  title={UnityVideo: Unified Multi-Modal Multi-Task Learning for Enhancing World-Aware Video Generation},
  author={Huang, Jiehui and Zhang, Yuechen and He, Xu and Gao, Yuan and Cen, Zhi and Xia, Bin and Zhou, Yan and Tao, Xin and Wan, Pengfei and Jia, Jiaya},
  journal={arXiv preprint arXiv:2512.07831},
  year={2025}
}

@article{blattmann2023stable,
  title={Stable video diffusion: Scaling latent video diffusion models to large datasets},
  author={Blattmann, Andreas and Dockhorn, Tim and Kulal, Sumith and Mendelevitch, Daniel and Kilian, Maciej and Lorenz, Dominik and Levi, Yam and English, Zion and Voleti, Vikram and Letts, Adam and others},
  journal={arXiv preprint arXiv:2311.15127},
  year={2023}
}

@inproceedings{voleti2024sv3d,
  title={Sv3d: Novel multi-view synthesis and 3d generation from a single image using latent video diffusion},
  author={Voleti, Vikram and Yao, Chun-Han and Boss, Mark and Letts, Adam and Pankratz, David and Tochilkin, Dmitry and Laforte, Christian and Rombach, Robin and Jampani, Varun},
  booktitle={European Conference on Computer Vision},
  pages={439--457},
  year={2024},
  organization={Springer}
}

@article{wu2025qwenimage,
  title={Qwen-image technical report},
  author={Wu, Chenfei and Li, Jiahao and Zhou, Jingren and Lin, Junyang and Gao, Kaiyuan and Yan, Kun and Yin, Sheng-ming and Bai, Shuai and Xu, Xiao and Chen, Yilei and others},
  journal={arXiv preprint arXiv:2508.02324},
  year={2025}
}

@article{cai2025zimage,
  title={Z-image: An efficient image generation foundation model with single-stream diffusion transformer},
  author={Cai, Huanqia and Cao, Sihan and Du, Ruoyi and Gao, Peng and Hoi, Steven and Hou, Zhaohui and Huang, Shijie and Jiang, Dengyang and Jin, Xin and Li, Liangchen and others},
  journal={arXiv preprint arXiv:2511.22699},
  year={2025}
}

@article{kong2024hunyuanvideo,
  title={Hunyuanvideo: A systematic framework for large video generative models},
  author={Kong, Weijie and Tian, Qi and Zhang, Zijian and Min, Rox and Dai, Zuozhuo and Zhou, Jin and Xiong, Jiangfeng and Li, Xin and Wu, Bo and Zhang, Jianwei and others},
  journal={arXiv preprint arXiv:2412.03603},
  year={2024}
}

@article{guo2023animatediff,
  title={Animatediff: Animate your personalized text-to-image diffusion models without specific tuning},
  author={Guo, Yuwei and Yang, Ceyuan and Rao, Anyi and Liang, Zhengyang and Wang, Yaohui and Qiao, Yu and Agrawala, Maneesh and Lin, Dahua and Dai, Bo},
  journal={arXiv preprint arXiv:2307.04725},
  year={2023}
}

@article{li2026skyreels,
  title={SkyReels-V3 Technique Report},
  author={Li, Debang and Fei, Zhengcong and Li, Tuanhui and Dou, Yikun and Chen, Zheng and Yang, Jiangping and Fan, Mingyuan and Xu, Jingtao and Wang, Jiahua and Gu, Baoxuan and others},
  journal={arXiv preprint arXiv:2601.17323},
  year={2026}
}

@inproceedings{zhang2023sadtalker,
  title={Sadtalker: Learning realistic 3d motion coefficients for stylized audio-driven single image talking face animation},
  author={Zhang, Wenxuan and Cun, Xiaodong and Wang, Xuan and Zhang, Yong and Shen, Xi and Guo, Yu and Shan, Ying and Wang, Fei},
  booktitle={Proceedings of the IEEE/CVF conference on computer vision and pattern recognition},
  pages={8652--8661},
  year={2023}
}

@article{wang2021audio2head,
  title={Audio2head: Audio-driven one-shot talking-head generation with natural head motion},
  author={Wang, Suzhen and Li, Lincheng and Ding, Yu and Fan, Changjie and Yu, Xin},
  journal={arXiv preprint arXiv:2107.09293},
  year={2021}
}

@inproceedings{dit,
  title={Scalable diffusion models with transformers},
  author={Peebles, William and Xie, Saining},
  booktitle={Proceedings of the IEEE/CVF international conference on computer vision},
  pages={4195--4205},
  year={2023}
}

@inproceedings{huang2024vbench,
  title={Vbench: Comprehensive benchmark suite for video generative models},
  author={Huang, Ziqi and He, Yinan and Yu, Jiashuo and Zhang, Fan and Si, Chenyang and Jiang, Yuming and Zhang, Yuanhan and Wu, Tianxing and Jin, Qingyang and Chanpaisit, Nattapol and others},
  booktitle={Proceedings of the IEEE/CVF Conference on Computer Vision and Pattern Recognition},
  pages={21807--21818},
  year={2024}
}

@inproceedings{liu2025hoigen,
  title={Hoigen-1m: A large-scale dataset for human-object interaction video generation},
  author={Liu, Kun and Liu, Qi and Liu, Xinchen and Li, Jie and Zhang, Yongdong and Luo, Jiebo and He, Xiaodong and Liu, Wu},
  booktitle={Proceedings of the Computer Vision and Pattern Recognition Conference},
  pages={24001--24010},
  year={2025}
}

@inproceedings{wang2025fantasytalking,
  title={Fantasytalking: Realistic talking portrait generation via coherent motion synthesis},
  author={Wang, Mengchao and Wang, Qiang and Jiang, Fan and Fan, Yaqi and Zhang, Yunpeng and Qi, Yonggang and Zhao, Kun and Xu, Mu},
  booktitle={Proceedings of the 33rd ACM International Conference on Multimedia},
  pages={9891--9900},
  year={2025}
}

@article{gan2025omniavatar,
  title={Omniavatar: Efficient audio-driven avatar video generation with adaptive body animation},
  author={Gan, Qijun and Yang, Ruizi and Zhu, Jianke and Xue, Shaofei and Hoi, Steven},
  journal={arXiv preprint arXiv:2506.18866},
  year={2025}
}

@inproceedings{yu2024gaussiantalker,
  title={Gaussiantalker: Speaker-specific talking head synthesis via 3d gaussian splatting},
  author={Yu, Hongyun and Qu, Zhan and Yu, Qihang and Chen, Jianchuan and Jiang, Zhonghua and Chen, Zhiwen and Zhang, Shengyu and Xu, Jimin and Wu, Fei and Lv, Chengfei and others},
  booktitle={Proceedings of the 32nd ACM International Conference on Multimedia},
  pages={3548--3557},
  year={2024}
}

@article{yang2025infinitetalk,
  title={Infinitetalk: Audio-driven video generation for sparse-frame video dubbing},
  author={Yang, Shaoshu and Kong, Zhe and Gao, Feng and Cheng, Meng and Liu, Xiangyu and Zhang, Yong and Kang, Zhuoliang and Luo, Wenhan and Cai, Xunliang and He, Ran and others},
  journal={arXiv preprint arXiv:2508.14033},
  year={2025}
}

@inproceedings{li2025anydressing,
  title={Anydressing: Customizable multi-garment virtual dressing via latent diffusion models},
  author={Li, Xinghui and Sun, Qichao and Zhang, Pengze and Ye, Fulong and Liao, Zhichao and Feng, Wanquan and Zhao, Songtao and He, Qian},
  booktitle={2025 IEEE/CVF Conference on Computer Vision and Pattern Recognition (CVPR)},
  pages={23723--23733},
  year={2025},
  organization={IEEE}
}

@article{gao2025wans2v,
  title={Wan-s2v: Audio-driven cinematic video generation},
  author={Gao, Xin and Hu, Li and Hu, Siqi and Huang, Mingyang and Ji, Chaonan and Meng, Dechao and Qi, Jinwei and Qiao, Penchong and Shen, Zhen and Song, Yafei and others},
  journal={arXiv preprint arXiv:2508.18621},
  year={2025}
}

@inproceedings{lin2025omnihuman,
  title={Omnihuman-1: Rethinking the scaling-up of one-stage conditioned human animation models},
  author={Lin, Gaojie and Jiang, Jianwen and Yang, Jiaqi and Zheng, Zerong and Liang, Chao and Zhang, Yuan and Liu, Jingtuo},
  booktitle={Proceedings of the IEEE/CVF International Conference on Computer Vision},
  pages={13847--13858},
  year={2025}
}

@article{shi2025diffmoe,
  title={Diffmoe: Dynamic token selection for scalable diffusion transformers},
  author={Shi, Minglei and Yuan, Ziyang and Yang, Haotian and Wang, Xintao and Zheng, Mingwu and Tao, Xin and Zhao, Wenliang and Zheng, Wenzhao and Zhou, Jie and Lu, Jiwen and others},
  journal={arXiv preprint arXiv:2503.14487},
  year={2025}
}

@article{ditmoe,
  title={Scaling diffusion transformers to 16 billion parameters},
  author={Fei, Zhengcong and Fan, Mingyuan and Yu, Changqian and Li, Debang and Huang, Junshi},
  journal={arXiv preprint arXiv:2407.11633},
  year={2024}
}

@inproceedings{prajwal2020lip,
  title={A lip sync expert is all you need for speech to lip generation in the wild},
  author={Prajwal, KR and Mukhopadhyay, Rudrabha and Namboodiri, Vinay P and Jawahar, CV},
  booktitle={Proceedings of the 28th ACM international conference on multimedia},
  pages={484--492},
  year={2020}
}

@article{sambody,
  title={SAM 3D Body: Robust Full-Body Human Mesh Recovery},
  author={Yang, Xitong and Kukreja, Devansh and Pinkus, Don and Sagar, Anushka and Fan, Taosha and Park, Jinhyung and Shin, Soyong and Cao, Jinkun and Liu, Jiawei and Ugrinovic, Nicolas and others},
  journal={arXiv preprint arXiv:2602.15989},
  year={2026}
}

@article{zhou2020makelttalk,
  title={Makelttalk: speaker-aware talking-head animation},
  author={Zhou, Yang and Han, Xintong and Shechtman, Eli and Echevarria, Jose and Kalogerakis, Evangelos and Li, Dingzeyu},
  journal={ACM Transactions On Graphics (TOG)},
  volume={39},
  number={6},
  pages={1--15},
  year={2020},
  publisher={ACM New York, NY, USA}
}

@article{education,
  title={Effectiveness of an avatar educational application for improving heart failure patients’ knowledge and self-care behaviors: A pragmatic randomized controlled trial},
  author={Wonggom, Parichat and Nolan, Paul and Clark, Robyn A and Barry, Tracey and Burdeniuk, Christine and Nesbitt, Katie and O'Toole, Kathryn and Du, Huiyun},
  journal={Journal of advanced nursing},
  volume={76},
  number={9},
  pages={2401--2415},
  year={2020},
  publisher={Wiley Online Library}
}

@inproceedings{huang2024makeyouranchor,
  title={Make-Your-Anchor: A Diffusion-based 2D Avatar Generation Framework},
  author={Huang, Ziyao and Tang, Fan and Zhang, Yong and Cun, Xiaodong and Cao, Juan and Li, Jintao and Lee, Tong-Yee},
  booktitle={Proceedings of the IEEE/CVF Conference on Computer Vision and Pattern Recognition (CVPR)},
  year={2024}
}

@article{education2,
  title={Effectiveness of avatar-based technology in patient education for improving chronic disease knowledge and self-care behavior: a systematic review},
  author={Wonggom, Parichat and Kourbelis, Constance and Newman, Peter and Du, Huiyun and Clark, Robyn A},
  journal={JBI evidence synthesis},
  volume={17},
  number={6},
  pages={1101--1129},
  year={2019},
  publisher={LWW}
}

@inproceedings{ke2021musiq,
  title={MUSIQ: Multi-scale image quality transformer},
  author={Ke, Junjie and Wang, Qifei and Wang, Yilin and Milanfar, Peyman and Yang, Feng},
  booktitle={Proceedings of the IEEE/CVF International Conference on Computer Vision},
  pages={5148--5157},
  year={2021}
}

@article{schuhmann2022laion,
  title={LAION-5B: An open large-scale dataset for training next generation image-text models},
  author={Schuhmann, Christoph and Beaumont, Romain and Vencu, Richard and Gordon, Cade and Wightman, Ross and Cherti, Mehdi and Coombes, Theo and Katta, Aarush and Mullis, Clayton and Wortsman, Mitchell and others},
  journal={Advances in Neural Information Processing Systems},
  volume={35},
  pages={25278--25294},
  year={2022}
}

@article{gemini2024,
  title={Gemini: A family of highly capable multimodal models},
  author={Gemini Team and Anil, Rohan and Borgeaud, Sebastian and Alayrac, Jean-Baptiste and Yu, Jiahui and Sorber, Radu and Schalkwyk, Johan and Dai, Andrew M and Hauth, Anja and Millican, Katie and others},
  journal={arXiv preprint arXiv:2312.11805},
  year={2024}
}

@article{lin2025cyberhost,
  title={CyberHost: Taming Audio-driven Avatar Diffusion Model with Region Codebook Attention},
  author={Lin, Gaojie and Zheng, Jianwen and Yang, Jiaqi and Zheng, Zerong and Liang, Chao and Zhang, Yuan and Liu, Jingtuo},
  journal={arXiv preprint arXiv:2409.01876},
  year={2025}
}

@article{oquab2024dinov2,
  title={DINOv2: Learning robust visual features without supervision},
  author={Oquab, Maxime and Darcet, Timoth{\'e}e and Moutakanni, Th{\'e}o and Vo, Huy and Szafraniec, Marc and Khalidov, Vasil and Fernandez, Pierre and Haziza, Daniel and Massa, Francisco and El-Nouby, Alaaeldin and others},
  journal={Transactions on Machine Learning Research},
  year={2024}
}

@article{xue2025mogan,
  title={MoGAN: Improving Motion Quality in Video Diffusion via Few-Step Motion Adversarial Post-Training},
  author={Xue, Haotian and Chen, Qi and Wang, Zhonghao and Huang, Xun and Shechtman, Eli and Xie, Jinrong and Chen, Yongxin},
  journal={arXiv preprint arXiv:2511.21592},
  year={2025}
}

@article{wu2025geometry,
  title={Geometry forcing: Marrying video diffusion and 3d representation for consistent world modeling},
  author={Wu, Haoyu and Wu, Diankun and He, Tianyu and Guo, Junliang and Ye, Yang and Duan, Yueqi and Bian, Jiang},
  journal={arXiv preprint arXiv:2507.07982},
  year={2025}
}

@inproceedings{yang2023dwpose,
  title={Effective whole-body pose estimation with two-stages distillation},
  author={Yang, Zhendong and Zeng, Ailing and Yuan, Chun and Li, Yu},
  booktitle={Proceedings of the IEEE/CVF International Conference on Computer Vision},
  pages={4210--4220},
  year={2023}
}

@inproceedings{shan2020understanding,
  title={Understanding human hands in contact at internet scale},
  author={Shan, Dandan and Geng, Jiaqi and Shu, Michelle and Fouhey, David F},
  booktitle={Proceedings of the IEEE/CVF conference on computer vision and pattern recognition},
  pages={9869--9878},
  year={2020}
}

@inproceedings{deng2019arcface,
  title={ArcFace: Additive angular margin loss for deep face recognition},
  author={Deng, Jiankang and Guo, Jia and Xue, Niannan and Zafeiriou, Stefanos},
  booktitle={Proceedings of the IEEE/CVF Conference on Computer Vision and Pattern Recognition},
  pages={4690--4699},
  year={2019}
}

@article{chefer2025videojam,
  title={Videojam: Joint appearance-motion representations for enhanced motion generation in video models},
  author={Chefer, Hila and Singer, Uriel and Zohar, Amit and Kirstain, Yuval and Polyak, Adam and Taigman, Yaniv and Wolf, Lior and Sheynin, Shelly},
  journal={arXiv preprint arXiv:2502.02492},
  year={2025}
}

@article{guo2026dreamidv,
  title={DreamID-V: Bridging the Image-to-Video Gap for High-Fidelity Face Swapping via Diffusion Transformer},
  author={Guo, Xu and Ye, Fulong and Li, Xinghui and Tu, Pengqi and Zhang, Pengze and Sun, Qichao and Zhao, Songtao and Hou, Xiangwang and He, Qian},
  journal={arXiv preprint arXiv:2601.01425},
  year={2026}
}

@article{guo2026dreamidomni,
  title={DreamID-Omni: Unified Framework for Controllable Human-Centric Audio-Video Generation},
  author={Guo, Xu and Ye, Fulong and Sun, Qichao and Chen, Liyang and Li, Bingchuan and Zhang, Pengze and Liu, Jiawei and Zhao, Songtao and He, Qian and Hou, Xiangwang},
  journal={arXiv preprint arXiv:2602.12160},
  year={2026}
}

@article{mediapipe,
  title={Mediapipe: A framework for building perception pipelines},
  author={Lugaresi, Camillo and Tang, Jiuqiang and Nash, Hadon and McClanahan, Chris and Uboweja, Esha and Hays, Michael and Zhang, Fan and Chang, Chuo-Ling and Yong, Ming Guang and Lee, Juhyun and others},
  journal={arXiv preprint arXiv:1906.08172},
  year={2019}
}

@inproceedings{chung2016syncnet,
  title={Out of time: Automated lip sync in the wild},
  author={Chung, Joon Son and Zisserman, Andrew},
  booktitle={Asian Conference on Computer Vision},
  pages={251--263},
  year={2016},
  organization={Springer}
}

@article{commerce,
  title={Avatar effect of AI-enabled virtual streamers on consumer purchase intention in e-commerce livestreaming},
  author={Sun, Luping and Tang, Yanfei},
  journal={Journal of Consumer Behaviour},
  volume={23},
  number={6},
  pages={2999--3010},
  year={2024},
  publisher={Wiley Online Library}
}

@article{zhou2026omnishow,
  title={OmniShow: Unifying Multimodal Conditions for Human-Object Interaction Video Generation},
  author={Zhou, Donghao and Liu, Guisheng and Yang, Hao and Li, Jiatong and Lin, Jingyu and Huang, Xiaohu and Liu, Yichen and Gao, Xin and Chen, Cunjian and Wen, Shilei and others},
  journal={arXiv preprint arXiv:2604.11804},
  year={2026}
}

@article{liao2025humanaesexpert,
  title={Humanaesexpert: Advancing a multi-modality foundation model for human image aesthetic assessment},
  author={Liao, Zhichao and Liu, Xiaokun and Qin, Wenyu and Li, Qingyu and Wang, Qiulin and Wan, Pengfei and Zhang, Di and Zeng, Long and Feng, Pingfa},
  journal={arXiv preprint arXiv:2503.23907},
  year={2025}
}

@article{framepack,
  title={Packing input frame context in next-frame prediction models for video generation},
  author={Zhang, Lvmin and Agrawala, Maneesh},
  journal={arXiv preprint arXiv:2504.12626},
  year={2025}
}
\end{document}